%% file: main.tex
\pdfoutput=1

\documentclass[11pt]{article}

\usepackage[]{EMNLP2022}

\usepackage{times}
\usepackage{latexsym}

\usepackage[T1]{fontenc}

\usepackage[utf8]{inputenc}

\usepackage{microtype}

\usepackage{inconsolata}

\usepackage{todonotes}
\usepackage{graphicx}
\usepackage{amsmath}
\usepackage{amssymb}
\usepackage{multirow}
\usepackage{booktabs}

\newcommand{\mypara}[1]{\noindent\textbf{#1}}
\newcommand\bfunder[1]{\textbf{\underline{#1}}}

\usepackage{enumitem}
\definecolor{ForestGreen}{rgb}{0.56, 0.74, 0.56}
\definecolor{OliveGreen}{rgb}{0.0, 0.26, 0.15}
\definecolor{BlueGreen}{rgb}{0.1, 0.7, 0.54}

\title{Pre-training Transformer Models with Sentence-Level Objectives for Answer Sentence Selection}

\author{Luca Di Liello$^{1}$\thanks{\ \ Work done as an intern at Amazon Alexa AI}\ , Siddhant Garg$^{2}$, Luca Soldaini$^{3}$\thanks{\ \ Work completed at Amazon Alexa AI}\ , Alessandro Moschitti$^{2}$\\
$^{1}$University of Trento , $^{2}$Amazon Alexa AI, $^{3}$Allen Institute for AI\\
\texttt{luca.diliello@unitn.it} \\ \texttt{\{sidgarg,amosch\}@amazon.com} \\
\texttt{lucas@allenai.org} \\
}

\begin{document}
\maketitle
\begin{abstract}
An important task for designing QA systems is answer sentence selection (AS2): selecting the sentence containing (or constituting) the answer to a question from a set of retrieved relevant documents.
In this paper, we propose three novel sentence-level transformer pre-training objectives that incorporate paragraph-level semantics within and across documents, to improve the performance of transformers for AS2, and mitigate the requirement of large labeled datasets.
Specifically, the model is tasked to predict whether: (i) two sentences are extracted from the same paragraph, (ii) a given sentence is extracted from a given paragraph, and (iii) two paragraphs are extracted from the same document.
Our experiments on three public and one industrial AS2 datasets demonstrate the empirical superiority of our pre-trained transformers over baseline models such as RoBERTa and ELECTRA for AS2. \end{abstract}

\section{Introduction}
Question Answering (QA) finds itself at the core of several commercial applications, for e.g., virtual
assistants such as Google Home, Alexa and Siri. Answer Sentence Selection (AS2) is an important task for QA Systems operating on unstructured text such as web documents. When presented with a set of relevant documents for a question (retrieved from a web index), AS2 aims to find the best answer sentence for the question. 

The recent popularity of pre-trained transformers~\cite{devlin-etal-2019-bert,liu2019roberta,clark2020electra}, has made them the de-facto approach for most QA tasks, including AS2. Several research works~\cite{garg2019tanda,laskar-etal-2020-contextualized,lauriola2021answer} fine-tune transformers for AS2, by posing it as a sentence-pair task and performing inference over the encoded representations of the question and answer candidates.

AS2 is a knowledge-intensive complex reasoning task, where the answer candidates for a question can stem from multiple documents, possibly on different topics linked to concepts in the question. 
While there have been recent works~\cite{ginzburg-etal-2021-self,caciularu-etal-2021-cdlm-cross} proposing pre-training strategies for obtaining multi-document aware document representations over long input encoders such as the Longformer~\cite{Beltagy2020Longformer}, there has been limited research~\cite{giorgi-etal-2021-declutr} on enhancing sentence-pair representations with paragraph and document level semantics.

Furthermore, obtaining high quality human labeled examples for AS2 is expensive and time consuming, due to the large number of answer candidates to be annotated for each question. Domain-specific AS2 datasets such as WikiQA~\cite{yang2015wikiqa} and TREC-QA~\cite{wang-etal-2007-jeopardy} only contain a few thousand questions. \citet{garg2019tanda} show that effectively fine-tuning pre-trained transformers on these domain specific AS2 datasets requires an intermediate fine-tuning transfer on a large scale AS2 dataset (ASNQ).

Towards improving the downstream performance of pre-trained transformers for AS2 and mitigating the requirement of large scale labeled data for fine-tuning, we propose three novel sentence-level transformer pre-training objectives, which can incorporate paragraph-level semantics across multiple documents. Analogous to the sentence-pair nature of AS2, we design our pre-training objectives to operate over a pair of input text sequences. The model is tasked with predicting: (i) whether the sequences are two sentences extracted from the same paragraph, (ii) whether the first sequence is a sentence that is extracted from the second sequence (paragraph), and (iii) whether the sequences are two paragraphs belonging to the same document.

We evaluate our paragraph-aware pre-trained transformers for AS2 on three popular public datasets: ASNQ, WikiQA and TREC-QA; and one industrial QA benchmark~\footnote{We will release code and pre-trained models at \url{https://github.com/amazon-research/
wqa-pretraining}}. Results show that our pre-training can improve the performance of fine-tuning baseline transformers such as RoBERTa and ELECTRA on AS2 by ${\sim}3{-}4\%$ points without requiring any additional data (labeled/unlabeled). 

\section{Related Work}

\mypara{Answer Sentence Selection (AS2)} Earlier approaches for AS2 used  CNNs~\cite{Severyn2015LearningTR} or alignment networks~\cite{Shen2017InterWeightedAN,Tran2018TheCA} to learn and score question and answer representations. Since then, compare-and-aggregate architectures have also been extensively studied~\cite{Wang2017ACM,Bian2017ACM,Yoon2019ACM}.~
~\citeauthor{garg2019tanda} achieved state-of-the-art results by fine-tuning transformers on a large QA corpora first, and then adapting to a smaller AS2 dataset.

\mypara{Token-Level Pre-training Objectives} Masked Language Modeling (MLM) is one of the most popular token-level pre-training objectives used for transformers~\cite{devlin-etal-2019-bert,liu2019roberta}. Some other models trained using token-level pre-training objectives are~\citet{yang2020xlnet} and \citet{clark2020electra}. ~\citet{joshi-etal-2020-spanbert} modify MLM to a span-prediction objective to make the model generalize well to machine reading tasks in QA.  

\mypara{Sentence-Level Pre-training Objectives} In addition to MLM, ~\citet{devlin-etal-2019-bert} uses the next sentence prediction (NSP) objective, which was later shown to not provide empirically improvements over MLM by~\citeauthor{liu2019roberta} (possibly due to the task being very simple). \citet{lan2020albert} propose a sentence order prediction (SOP) objective. \citet{ippolito-etal-2020-toward} enhance NSP to a multiple-choice prediction of the next sentence over a set of candidates, however they embed each sentence independently without cross-attention between them similar to~\cite{DBLP:journals/corr/abs-1908-10084}. \citet{gao-etal-2021-simcse} propose a supervised contrastive learning approach for enhancing sentence representations for textual similarity tasks.

\mypara{Paragraph/Document-level Semantics}
~\cite{DBLP:journals/corr/abs-1901-09128} pre-train Bi-HLSTMs for obtaining hierarchical document representations. HIBERT~\cite{zhang-etal-2019-hibert} uses document-level token masking and sentence masking pre-training objectives for generative tasks such as document summarization. Transformer pre-training objectives at different granularities of document semantics are discussed in~\cite{DBLP:journals/corr/abs-2012-14500} for fact verification, and in~\cite{Chang2020Pre-training} for retrieval. \citeauthor{ginzburg-etal-2021-self,caciularu-etal-2021-cdlm-cross} propose pre-training strategies for document embeddings for retrieval tasks such as document-matching. DeCLUTR~\cite{giorgi-etal-2021-declutr} uses contrastive learning for cross-encoding two sentences coming from the same/different documents in a transformer, and is evaluated on pairwise binary classification tasks like natural language inference. Our work differs from this since we use a cross-encoder architecture to capture cross-attention between the question and answer, and evaluate our approach on the relevance ranking task of AS2 over hundreds of candidates. Contemporary works~\cite{di-liello-etal-2022-paragraph} pre-train transformers using paragraph-aware objectives for multi-sentence inference tasks. Our work differs from this since we only encode a pair of sentences using the transformer, while the former encode multiple sentences and use sophisticated prediction heads to aggregate information across multiple representations. 

\mypara{Transformers for Long Inputs}
Longformer~\cite{Beltagy2020Longformer}, Big Bird~\cite{zaheer2020bigbird}, etc. model very long inputs (e.g, entire documents) by reducing the complexity of transformer attention. This provides longer context, which is useful for machine reading and summarization.

\section{Answer Sentence Selection (AS2)}
In this section we formally define the task of AS2. Given a question $q$ and a set of answer candidates $A{=}\{a_1,{\dots},a_n\}$, the objective is to select the candidate $\bar{a} \ {\in} \ A$ that best answers $q$. AS2 can be modeled as a ranking task over $A$ to learn a scoring function $f: Q {\times} A \ {\rightarrow} \ \mathbb{R}$ that predicts the probability $f(q,a)$ of an answer candidate $a$ being correct. The best answer $\bar{a}$ corresponds to $\text{argmax}^n_{i=1} \text{ } f(q, a_i)$. Pre-trained transformers are used as QA pair encoders for AS2 to approximate the function $f$.

\section{Sentence-Level Pre-training Objectives}
\label{sec:objectives}

Documents are typically organized into paragraphs, by humans, to address the document's general topic from different viewpoints. We propose three pre-training objectives to exploit the intrinsic information contained in the structure of documents. For all these objectives, we provide a pair of text sequences as input to the transformer to jointly reason over them, analogous to the AS2 task.

\paragraph{Spans in Same Paragraph (SSP)} 
Given two sequences $(A, B)$ as input to the transformer, the objective is to predict if $A$ and $B$ belong to the same paragraph in a document. To create positive pairs $(A, B)$, given a document $D$, we extract two small, contiguous and disjoint subsets of sentences to be used as $A$ and $B$ from a single paragraph $P_i \in D$. To create negative pairs, we sample spans of sentences $B'$ from different paragraphs $P_j, j \ne i$ in the same document $D$ (hard negatives) and also from different documents (easy negatives). The negative pairs correspond to $(A, B')$. 
Posing the above pre-training objective in terms of spans (instead of sentences) allows for modifying the lengths of the inputs $A, B$ (by changing number of sentences ${\in} A, B$).
When fine-tuning transformers for AS2, typically the question is provided as the first input and a \emph{longer} answer candidate/paragraph is provided as the second input. For our experiments (Sec~\ref{sec:experiments}), we use a longer span for input B than A.

\paragraph{Span in Paragraph (SP)}
Given two sequences $(A,B)$ as input to the transformer, the objective is to predict if $A$ is a span of text extracted from a paragraph $B$ in a document. To create positive pairs $(A, B)$, given a paragraph $P_i$ in a document $D$, we extract a small contiguous span of sentences $A$ from it and create the input pair as $(A, P_i \setminus A)$. To create negative pairs, we select other paragraphs $P_j, j \ne i$ in the same document $D$ and remove a randomly chosen span $A'$ from each of them. The negative pairs correspond to $(A,P_j \setminus A')$. This is necessary to ensure that the model does not simply recognize whether the second input is a complete paragraph or a clipped version. To create easy negatives, we use the above approach for paragraphs $P_j$ sampled from documents other than $D$. 

\paragraph{Paragraphs in Same Document (PSD)}
Given two sequences $(A,B)$ as input to the transformer, the objective is to predict if $A$ and $B$ are paragraphs belonging to the same document. To create positive pairs $(A, B)$, given a document $D_k$, we randomly select paragraphs $P_i,P_j \in D_k$ and obtain a pair ${(P_i, P_j)}$. To create negative pairs, we randomly select $P_j' \notin D_k$, and obtain a pair $(P_i, P_j')$.

\section{Experiments}
\label{sec:experiments}

\subsection{Datasets}

\paragraph{Pre-training}
To eliminate any improvements stemming from the usage of more data, we perform pre-training on the same corpora as RoBERTa: English Wikipedia, the BookCorpus, OpenWebText and CC-News. 
We perform continuous pre-training starting from RoBERTa~\cite{liu2019roberta} and ELECTRA~\cite{clark2020electra} checkpoints, using a combination of our objectives with the original ones (MLM for RoBERTa and MLM + Token Detection for ELECTRA). Refer to Appendix~\ref{app:datasets} for complete details.

\paragraph{AS2 Fine-tuning}
We consider three public and one industrial AS2 benchmark as fine-tuning datasets for AS2 (statistics presented in Appendix~\ref{app:datasets}). We use standard evaluation metrics for AS2: Mean Average Precision (MAP), Mean Reciprocal Recall (MRR) and Precision@1 (P@1).

\input{Tables/AS2_table}

\begin{itemize}[wide, labelindent=0pt]
\itemsep-0.25em
    \item \textbf{ASNQ} is a large-scale AS2 dataset~\cite{garg2019tanda} with questions from Google search engine queries, and answer candidates extracted from a Wikipedia page. ASNQ is a modified version of the Natural Questions (NQ)~\cite{kwiatkowski-etal-2019-natural}, obtained by labeling sentences from long answers that contain the short answer as positives and all others as negatives. We use the dev and test splits released by ~\citeauthor{soldaini-moschitti-2020-cascade}\footnote{\url{https://github.com/alexa/wqa-cascade-transformers}}.

    \item \textbf{WikiQA} is a popular AS2 dataset~\cite{yang2015wikiqa} where questions are derived from query logs of the Bing search engine, and the answer candidates are extracted from a Wikipedia page. This dataset has a subset of questions having no correct answers (\textit{all-}) or having only correct answers (\textit{all+}). We remove both the (\textit{all-}) and (\textit{all+}) questions for our experiments (standard ``clean'' setting).

    \item \textbf{TREC-QA} is a popular AS2 dataset~\cite{wang-etal-2007-jeopardy} of factoid questions, extracted from the TREC-8 to TREC-13 QA tracks. The answer candidates are sentences that contain one or more non-stopwords in common with the question, extracted from multiple documents. 
    For the dev and test sets, we remove questions without answers, or having only correct or only incorrect answer candidates (``clean'' setting~\cite{shen-etal-2017-inter}).

    \item \textbf{WQA} A large scale industrial AS2 dataset containing \emph{non-representative de-identified} user questions from Alexa virtual assistant. For every question, ${\sim}15$ answer candidates are collected from a large web index of more than 100M documents using Elasticsearch. Results on WQA are presented relative to the RoBERTa-Base baseline due to the data being internal.

\end{itemize}

\subsection{Experimental Setup and Details}

We use our 3 pre-training objectives: SSP, SP and PSD, for both RoBERTa and ELECTRA, obtaining 6 different continuously pre-trained models.
We set the maximum pre-training steps to $400$k for SSP and $200$k for SP and PSD. This corresponds to each model processing ${\sim}210B$ tokens during pre-training, which is about $10\%$ of the ${\sim}2100B$ tokens used for pre-training RoBERTa. Notice also that the compute FLOPs are even less than the $10\%$ of the original training because we used a shorted max sequence length. More details about the continuous pre-training hyper-parameters are given in Appendix \ref{app:experiments}.

We also combine all 3 objectives together (SSP+SP+PSD) for both RoBERTa and ELECTRA, with the same setting as SSP. We fine-tune each of our pre-trained models on all four AS2 datasets (with early stopping on the dev set) and compute results on their respective test splits. 

\mypara{Baselines} We use RoBERTa and ELECTRA models as baselines. We also use TANDA~\cite{garg2019tanda}, the state of the art for AS2, as an upper-bound baseline as it uses an additional intermediate transfer step on ASNQ (${\sim}20M$ labeled QA pairs). Note that we don't consider ~\citeauthor{ginzburg-etal-2021-self,caciularu-etal-2021-cdlm-cross, Chang2020Pre-training} as baselines as they are designed for document-matching and retrieval tasks, and ~\citeauthor{Beltagy2020Longformer,zaheer2020bigbird} as they are used for long-context tasks like MR and summarization.

\subsection{Results}

We present results of our pre-trained models on the AS2 datasets in Table~\ref{tab:results_as2}. We observe that the models trained with our pre-training objectives significantly outperform the baseline models when fine-tuned for the AS2 tasks. For example, on ASNQ, using our SP objective with RoBERTa-Base gains $2.3\%$ in P@1 over the baseline RoBERTa-Base model. On WikiQA, the performance gap is even larger with the SSP objective corresponding to $4.6\%$ points for RoBERTa-Base and $5.4\%$ for ELECTRA-Base over the corresponding baselines. Performance improvements on TREC-QA and WQA are smaller but consistent, around $1\%$ and $0.6\%$ in P@1. Combining SSP+SP+PSD together consistently achieves either the best results (TREC-QA and WQA), or close to the best results (ASNQ and WikiQA).

For questions in ASNQ and WikiQA, all candidate answers are extracted from a \emph{single} Wikipedia document, while for TREC-QA and WQA, candidate answers come from \emph{multiple} documents extracted from heterogeneous web sources. By design of our objectives SSP, SP and PSD, they perform differently when fine-tuning on different datasets. For example, SSP aligns well with ASNQ and WikiQA as they contain many negative candidates, per question, extracted from the same document as the positive (i.e, `hard' negatives). As per our design of the SSP objective, for every positive sequence pair, we sample 2 `hard' negatives coming from the same document as the positive pair. The presence of hard negatives is of particular importance for WikiQA and ASNQ, as it forces the models to learn and contrast more subtle differences between answer candidates, which might likely be more related as they come from the same document.

On the other hand, PSD is designed so as to see paragraphs from same or different documents (with no analogous concept of `hard' negatives of SSP and SP). For this reason, PSD is better aligned for fine-tuning on datasets where candidates are extracted from multiple documents, such as WQA and TREC-QA.

\paragraph{Comparison with TANDA} For RoBERTa, our pre-trained models can surprisingly improve/achieve comparable performance to TANDA. Note that our models achieve this performance without using the latter's additional ${\sim}20M$ labeled ASNQ QA pairs. This lends support to our pre-training objectives mitigating the requirement of large scale labeled data for AS2 fine-tuning. For ELECTRA, we only observe comparable performance to TANDA for WQA and TREC-QA.

\input{Tables/AS2_only_mlm_table_small}
\input{Tables/pretraining_table}

\paragraph{Ablation: MLM-only Pre-training}
To mitigate any improvements stemming from the specific data sampling techniques used by our objectives, we pre-train 3 models (starting from RoBERTa-Base) with the same data sampling as each of the SSP, SP and PSD models, but only using the MLM objective.

We report results in Table~\ref{tab:results_as2_only_mlm_small}, and observe that, almost always, models pre-trained only with MLM under-perform models trained with SSP, SP and PSD objectives in addition to MLM. Thus, the empirical improvements of our methods are derived from the novel pre-training objectives, and not data sampling. Surprisingly, for some models, the MLM-only continuous pre-training performs worse than the baseline RoBERTa-Base. We believe that restarting the training with a different learning-rate\footnote{The original models use a triangular learning-rate}, a shorter sequence length, and without the original optimizer and scheduler internal states (for a small amount of steps) is sub-optimal for the model.

\paragraph{Ablation: Pre-training Task `Difficulty'}
We evaluate the pre-trained models (after convergence) on their specific tasks over the validation split of Wikipedia (to enable evaluating baselines such as BERT and ALBERT). Table~\ref{tab:results_pretraining} summarizes the accuracy and F1 of the models on the various tasks.

The results show that our objectives are generally \emph{harder} than NSP (Next Sentence Prediction by ~\citealp{devlin-etal-2019-bert}) and SOP (Sentence Order Prediction by~\citealp{lan2020albert}). In fact, NSP and SOP have been shown to not add any significant performance improvements in addition to MLM \citep{liu2019roberta}, and this corresponds to the model being able to perform this task extremely well (dev accuracy $\sim94\%$ with NSP and $\sim97\%$ with SOP) without learning any new semantics that may be useful for downstream tasks.

On the other hand, our pre-training objectives are ``more challenging'' than these previously proposed objectives due to the requirement of reasoning over multiple paragraphs and multiple documents, addressing same or different topics at the same time. In fact, Table~\ref{tab:results_pretraining} shows that after convergence, our pre-trained model still finds it \emph{difficult} to achieve a higher accuracy for our sentence level pre-training tasks. Empirically in Table~\ref{tab:results_as2}, we observed that pre-training with our objectives is able to rank the more relevant answers at the top, which we hypothesize is due to the model learning how to reason over multiple paragraphs and documents already while performing continuous pre-training.

\section{Conclusion}
In this paper we have presented three sentence-level pre-training objectives for transformers to incorporate paragraph and document-level semantics. 
Our objectives predict whether (i) two sequences are sentences extracted from the same paragraph, (ii) first sequence is a sentence extracted from the second, and (iii) two sequences are paragraphs belonging to the same document.
We evaluate our pre-trained models for the task of AS2 on four datasets. Our results show that our pre-trained models outperform the baseline transformers such as RoBERTa and ELECTRA. 

\section*{Limitations}
\label{sec:limitations}

We only consider English language datasets for our experiments in this paper. However we hypothesize that our pre-training objectives should provide similar performance improvements when extended to other languages with limited morphology, like English. The pre-training objectives proposed in our work are designed considering Answer Sentence Selection (AS2) as the target task, and can be extended for other tasks like Natural Language Inference, Question-Question Similarity, etc. in future work.  The pre-training experiments in our paper require large amounts of GPU and compute resources (multiple NVIDIA A100 GPUs running for several days) to finish the model pre-training. This makes re-training models using our pre-training approaches computationally expensive using newer data. To mitigate this, we are releasing our code and pre-trained model checkpoints at \url{https://github.com/amazon-research/
wqa-pretraining}, which can directly be used by fine-tuning them on AS2 datasets. 

\section*{Acknowledgements}
We thank the anonymous reviewers and the ARR action-editor for their valuable suggestions. We would like to thank Thuy Vu for developing and sharing the WQA dataset.


\bibliographystyle{acl_natbib}
\bibliography{anthology,custom}

\appendix
\input{appendix}

\end{document}

%% file: Tables/AS2_table.tex
\begin{table*}[t!]
    \centering
    \resizebox{\linewidth}{!}{
    \begin{tabular}{lccccccccccccccc}
    \toprule
    \multirow{2}{*}{\textbf{Model}} & \multicolumn{3}{c}{\textbf{ASNQ}} & &\multicolumn{3}{c}{\textbf{WikiQA}} & & \multicolumn{3}{c}{\textbf{TREC-QA}} & & \multicolumn{3}{c}{\textbf{WQA}} \\
    \cmidrule{2-4} \cmidrule{6-8} \cmidrule{10-12} \cmidrule{14-16}
                           & \textbf{P@1}    & \textbf{MAP}    & \textbf{MRR}   & &\textbf{ P@1}     & \textbf{MAP}     & \textbf{MRR}   & & \textbf{P@1}   & \textbf{MAP}     & \textbf{MRR}  & & \textbf{P@1}    & \textbf{MAP}    & \textbf{MRR}    \\
    \midrule
    
    RoBERTa-Base     & 61.8 {\small (0.2)} & 66.9 {\small (0.1)} & 73.1 {\small (0.1)}                                      & & 78.3 {\small (2.8)} & 85.8 {\small (1.3)} & 87.2 {\small (1.3)}                                        & & 90.0 {\small (1.9)}  & 89.7 {\small (0.7)} & 94.4 {\small (1.1)}                               & & \multicolumn{3}{c}{Baseline}  \\
    {\small (\textbf{Ours})} RoBERTa + SSP       & \bfunder{64.1} {\small (0.3)} & \underline{68.1} {\small (0.2)} & \bfunder{74.5} {\small (0.3)}      & & \bfunder{82.9} {\small (0.7)} & \bfunder{88.7} {\small (0.3)} & \bfunder{89.9} {\small (0.4)}          & & 88.5 {\small (1.2)}  & 89.3 {\small (0.7)} & 93.6 {\small (0.6)}                               & & +0.2\% & \underline{+0.6\%} & +0.3\% \\
    {\small (\textbf{Ours})} RoBERTa + SP      & \bfunder{64.1} {\small (0.2)} & \bfunder{68.3} {\small (0.1)} & \bfunder{74.5} {\small (0.2)}        & & \underline{81.0} {\small (0.8)} & \underline{87.7} {\small (0.3)} & \underline{88.9} {\small (0.4)}    & & \textbf{90.9} {\small (2.6)} & 90.1 {\small (0.8)} & 94.7 {\small (1.3)}                       & & \underline{+0.4\%} & \underline{+0.7\%}  & \underline{+0.5\%} \\
    {\small (\textbf{Ours})} RoBERTa + PSD      & 62.6 {\small (0.4)} & \underline{67.7} {\small (0.2)} & \underline{73.7} {\small (0.3)}              & & 80.5 {\small (1.6)} & 86.4 {\small (1.1)} & 88.0 {\small (1.0)}                                        & & 90.3 {\small (1.3)} & \textbf{90.3} {\small (0.5)} & \textbf{95.1} {\small (0.7)}              & & \underline{+0.4\%}  & \underline{+0.7\%} & \underline{+0.5\%} \\
    
    {\small (\textbf{Ours})} RoBERTa + All      & \underline{63.9} {\small (0.4)} & \underline{68.0} {\small (0.1)} & \underline{74.1} {\small (0.2)}              & & \underline{82.5} {\small (0.9)} & \underline{88.2} {\small (0.4)} & \underline{89.5} {\small (0.4)}                                        & & 87.9 {\small (1.2)} & 89.3 {\small (0.7)} & 93.4 {\small (0.6)}              & & \bfunder{+0.5\%}  & \bfunder{+0.8\%} & \bfunder{+0.6\%} \\
    TANDA RoBERTa         & -     & -     & -     & & 83.0 {\small (1.3)}    & 88.5 {\small (0.8)}   & 89.9 {\small (0.8)}  & & 89.7 {\small (0.0)}    & 90.1 {\small (0.6)} & 94.1 {\small (0.4)}  & & +0.5\%   & +0.5\%    & +0.5\% \\
    
    \midrule

    ELECTRA-Base    & 62.4 {\small (0.4)} & 67.5 {\small (0.2)} & 73.6 {\small (0.2)}                                      & & 77.1 {\small (4.0)} & 85.0 {\small (2.6)} & 86.5 {\small (2.7)}                                        & & 90.3 {\small (1.7)} & 89.9 {\small (0.4)} & 94.0 {\small (0.9)}                                & & +1.0\%  & +1.2\% & +0.9\% \\
    {\small (\textbf{Ours})} ELECTRA + SSP       & \bfunder{65.3} {\small (0.3)} & \bfunder{69.7} {\small (0.2)} & \bfunder{75.7} {\small (0.2)}        & & \bfunder{82.5} {\small (2.0)} & \bfunder{88.6} {\small (1.4)} & \bfunder{90.0} {\small (1.4)}          & & 88.5 {\small (1.9)} & 89.6 {\small (0.7)} & 93.5 {\small (0.9)}                                & & \underline{+1.4\%} & \underline{+1.5\%} & \underline{+1.3\%} \\
    {\small (\textbf{Ours})} ELECTRA + SP      & \underline{65.0} {\small (0.2)} & \underline{69.0} {\small (0.1)} & \underline{75.1} {\small (0.1)}  & & \underline{81.8} {\small (2.3)} & \underline{88.1} {\small (1.5)} & \underline{89.5} {\small (1.5)}    & & 91.2 {\small (1.5)} & 90.3 {\small (0.7)} & 94.6 {\small (0.7)}     & & \underline{+1.4\%} & \underline{+1.5\%} & \underline{+1.3\%} \\
    {\small (\textbf{Ours})} ELECTRA + PSD      & \bfunder{65.3} {\small (0.4)} & \underline{68.9} {\small (0.3)} & \underline{75.1} {\small (0.3)}    & & 78.6 {\small (0.7)} & 85.6 {\small (0.7)} & 87.3 {\small (0.6)}                                        & & 85.9 {\small (2.2)} & 87.9 {\small (1.1)} & 92.2 {\small (1.1)}                                & & \bfunder{+1.6\%}  & \bfunder{+1.6\%} & \underline{+1.3\%} \\
    {\small (\textbf{Ours})} ELECTRA + All      & \underline{65.0} {\small (0.3)} & \underline{69.3} {\small (0.2)} & \underline{75.2} {\small (0.2)}    & & 80.8 {\small (1.9)} & 87.3 {\small (1.2)} & \underline{88.7} {\small (1.1)}                                        & & \bfunder{92.6} {\small (1.8)} & \textbf{90.4} {\small (0.4)} & \textbf{95.5} {\small (1.0)}                                & & \underline{+1.5\%}  & \bfunder{+1.6\%} & \bfunder{+1.4\%} \\
    TANDA ELECTRA         & -     & -     & -     & & 85.6 {\small (1.1)}    & 90.2 {\small (0.8)}   & 91.4 {\small (0.7)}  & & 92.6 {\small (1.5)}    & 91.6 {\small (0.7)} & 95.5 {\small (0.7)}  & & +1.9\%   & +1.6\%   & +1.5\% \\
    \bottomrule

    \end{tabular}
    }
    \caption{Results (with std. dev. across 5 runs in parentheses) of our pretrained transformers when fine-tuned on AS2 datasets. SSP, SP, PSD denote our pretraining objectives, and `All' denotes using SSP+SP+PSD together. TANDA uses \textbf{additional labeled data} as an intermediate transfer step. We underline statistically significant improvements over the baseline (T-test at a 95\% confidence level). Results on WQA are relative to the RoBERTa baseline.}
    \label{tab:results_as2}
    \vspace{0.5em}
\end{table*}

%% file: Tables/AS2_only_mlm_table_small.tex
\begin{table}[t!]
    \vspace{1em}
    \centering
    \resizebox{\linewidth}{!}{
    \begin{tabular}{lcccc}
    \toprule
    \textbf{Model+Data Sampling} & \textbf{ASNQ} & \textbf{WikiQA} & \textbf{TREC-QA} & \textbf{WQA} \\

    \midrule

    RoBERTa-Base                                   & 61.8                & 78.3               & 90.0  & Baseline \\

    \midrule

    + SSP Data (\textbf{MLM-only})        & 63.4               & 76.7               & 87.4            & -0.6\% \\
    + SSP                & \bfunder{64.1}     & \bfunder{82.9}     & \textbf{88.5}   & \bfunder{+0.2}\% \\

    \midrule

    + SP Data (\textbf{MLM-only})        & 62.8               & 76.8              & 88.8             & -1.0\% \\
    + SP                 & \bfunder{64.1}     & \bfunder{81.0}     & \textbf{90.9}   & \bfunder{+0.4}\% \\

    \midrule

    + PSD Data (\textbf{MLM-only})        & \bfunder{64.1}     & 79.1               & 87.1             & -1.3\% \\
    + PSD                & 62.6               & \bfunder{80.5}     & \bfunder{90.3}   & \bfunder{+0.4}\% \\

    \bottomrule

    \end{tabular}
    }
    \caption{P@1 of our pretrained models using SSP, SP and PSD objectives in addition to only MLM. We highlight in bold and underline results like in Table \ref{tab:results_as2}.}
    \label{tab:results_as2_only_mlm_small}
\end{table}

%% file: Tables/pretraining_table.tex
\begin{table}[t!]
    \vspace{1.5em}
    \centering
    \resizebox{\linewidth}{!}{
    \begin{tabular}{lccc}
    \toprule
    \textbf{Model + Pre-training Objective} & \textbf{Accuracy} & \textbf{F1} \\
    \midrule
    RoBERTa-Base + \textbf{SSP}      & 91.8   & 83.1 \\
    ELECTRA-Base + \textbf{SSP}      & 90.4   & 79.9 \\
    \midrule
    RoBERTa-Base + \textbf{SP}     & 91.3   & 83.3 \\
    ELECTRA-Base + \textbf{SP}     & 89.9   & 80.1 \\
    \midrule
    RoBERTa-Base + \textbf{PSD}     & 83.5   & 61.4 \\
    ELECTRA-Base + \textbf{PSD}     & 82.3   & 57.1 \\
    \midrule
    BERT~\cite{devlin-etal-2019-bert} (NSP) & 96.9 & 97.1\\
    ALBERT~\cite{lan2020albert} (SOP) & 93.7 & 94.7\\
    \bottomrule
    \end{tabular}
    }
    \caption{Comparison of accuracy and F1-score of pre-training objectives on the pre-training validation set.}
    \label{tab:results_pretraining}
\end{table}

%% file: appendix.tex
\clearpage

\section*{Appendix}

\section{Datasets}
\label{app:datasets}

\subsection{Pre-training}
For continued pre-training, we pre-process the English Wikipedia \footnote{\url{https://dumps.wikimedia.org/enwiki/20211101/}}, the BookCorpus\footnote{\url{https://huggingface.co/datasets/bookcorpusopen}}, OpenWebText \citep{Gokaslan2019OpenWeb} and the CC-News\footnote{\url{https://commoncrawl.org/2016/10/news-dataset-available/}} datasets. We do not use the STORIES dataset as it is no longer available for research use \footnote{\url{https://github.com/tensorflow/models/tree/archive/research/lm\_commonsense\#1-download-data-files}}. We clean every dataset by removing headers, titles, tables and any HTML content. For every document, we keep paragraphs containing at least 60 characters and documents containing at least 200 characters. After cleaning, we obtain 5GB, 10GB, 34GB and 360GB of raw text from the BookCorpus, Wikipedia, OpenWebText and CC-News respectively. We split paragraph into lists of sentences using the blingfire tokenizer\footnote{\url{https://github.com/microsoft/BlingFire}}. We present the details of our pre-training objectives in Section~\ref{sec:objectives}. We present the details on sampling lengths and number of negatives for each of the objectives below:

\begin{itemize}[wide, labelindent=0pt]
\itemsep-0.25em
    \item \textbf{Spans in Same Paragraph (SSP)}
    We randomly sample the number of sentences in $A$ in the interval $\left[1, 3\right]$ and $B$ in $\left[1, 5\right]$. This is to keep the inputs to the model analogous to those in AS2 (shorter question text, followed by longer answer text). We sample up to 2 hard negatives from the same paragraph as $A$ (if possible), and sample easy negatives from other documents so as to make the total number of negatives to be 4.
    \item \textbf{Span in Paragraph (SP)}
    We randomly sample the number of sentences in $A \in P_i$ in the interval $\left[1, 3\right]$. The number of sentences in the right part is given by the length of $P_i \setminus A$ (positive pair) or $P_j \setminus X_j$ (negative pair). Similar to SSP, we sample up to 2 hard negatives from the same document (if possible), and sample easy negatives from other documents so as to make the total number to be 4.
    \item \textbf{Paragraphs in Same Document (PSD)}
    We chose a random pair of paragraphs $A$ and $B$ from a single document and then we randomly sample 4 paragraphs from other documents to create the negative pairs with $A$.
\end{itemize}

\subsection{Fine-tuning}
Here we present statistics and links for downloading the AS2 datasets used: ASNQ\footnote{\url{https://github.com/alexa/wqa_tanda}}, WikiQA\footnote{\url{http://aka.ms/WikiQA}}, TREC-QA and WQA; to benchmark our pre-trained models. Table \ref{tab:as2_datasets} shows the number of unique questions and answer candidates for each dataset and for each split.

\begin{table}[ht!]
\centering
\small
\begin{tabular}{llccc}
\hline
    \textbf{Dataset} & \textbf{Split} & \textbf{\# Q} & \textbf{\# C} & \textbf{Avg. \# C/Q} \\
    
    \hline
    \multirow{3}{*}{ASNQ}       & Train   & 57,242 & 20,377,568 & 356.0 \\
                                & Dev     & 1,336  & 463,914    & 347.2 \\
                                & Test    & 1,336  & 466,148    & 348.9 \\

    \hline
    \multirow{3}{*}{WikiQA}     & Train & 2,118 & 20,360  & 9.6 \\
                                & Dev     & 122   & 1,126   & 9.2 \\
                                & Test    & 237   & 2,341   & 9.9 \\
    \hline

    \multirow{3}{*}{TREC-QA}    & Train   & 1,226 & 53,417     & 43.6 \\
                                & Dev     & 69    & 1,343      & 19.5 \\
                                & Test    & 68    & 1,442      & 21.2 \\

    \hline
    \multirow{3}{*}{WQA}        & Train   & 9,984     & 149,513     & 15.0 \\
                                & Dev     & 5,000     & 74,805      & 15.0 \\
                                & Test    & 5,000     & 74,712      & 14.9 \\
    \hline

\end{tabular}
\caption{Data Statistics for AS2 dataset. ``Avg. \# C/Q'' is the average number of answer candidates per question.}
\label{tab:as2_datasets}
\end{table}



\section{Experimental Setup}
\label{app:experiments}

We experiment with the \textit{base} architecture, which uses an hidden size of $768$, $12$ transformer layers, $12$ attention heads and feed-forward size of $3072$.

\paragraph{Pre-training} We perform continued pre-training starting from the publicly released checkpoints of RoBERTa-Base \cite{liu2019roberta} and ELECTRA-Base \cite{clark2020electra}. We optimize using Adam, which we instantiate with $\beta_1 = 0.9$, $\beta_2 = 0.999$ and $\epsilon = 10^{-8}$. We use a triangular learning rate with $10$k warmup steps. The peak learning rate is set to $1*10^{-4}$. We apply a weight decay of $0.01$, gradient clipping when values are larger than $1.0$ and dropout ratio is set to $0.1$. We set the batch size to $4096$ examples for every combination of models and objectives. We truncate the input sequences to $128$ tokens for SSP and to $256$ tokens with SP and PSD. Finally, we perform $400$k training steps with models using SSP and $200$k steps with the other objectives: SP and PSD. The total amount of tokens seen in the continued pre-training is the same for all models and equal to ${\sim}210B$. 

We combine the binary classification loss of SSP, SP and PSD with MLM for RoBERTa and with MLM (of the generator) and TD (token detection) for ELECTRA. For RoBERTa, we perform binary classification on the first {\small [CLS]} token in addition to MLM. For ELECTRA, using the generator + discriminator architecture, we perform MLM on the generator; and token-detection along with binary classification on the discriminator using our pre-training objectives. Through experimentation, for RoBERTa, we use equal weights for MLM and our pre-training objectives. For ELECTRA, we combine MLM, TD and our pre-training objectives with the weights 1.0, 50.0 and 1.0 respectively.

\paragraph{Fine-tuning} The evaluation of the models is performed on four different datasets for Answer Sentence Selection. We maintain the same hyper-parameters used in pre-training apart from the learning rate, the number of warmup steps and the batch size. We do early stopping on the development set if the number of non-improving validations (patience) is higher than 5. For ASNQ, we found that using a very large batch size is beneficial, providing a higher accuracy. We use a batch size of $2048$ examples on ASNQ for RoBERTa models and $1024$ for ELECTRA models. The peak learning rate is set to $1*10^{-5}$ for all models, and the number of warmup steps to $1000$. For WikiQA, TREC-QA and WQA, we select the best batch size out of $\{ 16, 32, 64 \}$ and learning rate out of $\{ 2*10^{-6}, 5*10^{-5}, 1*10^{-5}, 2*10^{-5} \}$ using cross-validation. We train the model for 6 epochs on ASNQ, and up to 40 epochs on WikiQA, TREC-QA, and WQA. The performance of practical AS2 systems is typically measured using Precision-at-1 P@1~\cite{garg-moschitti-2021-will}. In addition to P@1, we also use Mean Average Precision (MAP) and Mean Reciprocal Recall (MRR) to evaluate the ranking of the set of candidates produced by the model.

We used metrics from Torchmetrics~\cite{torchmetrics} to compute MAP, MRR, Precision@1 and Accuracy.

\section{Experiments and Results} 

\subsection{Ablation: MLM-only Pre-training}
Table~\ref{tab:results_as2_only_mlm} presents a more detailed comparison between models continuously pre-trained only with MLM and models using also the sentence-level classification loss functions we proposed in this paper.

\input{Tables/AS2_only_mlm_table}

\section{Qualitative Examples from AS2}
\label{app:qualitative}

We present some qualitative examples from the three public AS2 datasets. We highlight cases in which the baseline RoBERTa-Base model is unable to rank the correct answer in the top position, but where our model pretrained with SP is successful. The examples are provided in Table~\ref{tab:qualitative}.

\begin{table*}[t!]
    \centering
    \resizebox{\linewidth}{!}{
    \begin{tabular}{l}
    \toprule
    \textbf{ASNQ} \\
    \midrule

    \textbf{Q:} \textbf{how many players in football hall of fame}\\
    \textbf{A1:} {\color{red} Two coaches ( Marv Levy , Bud Grant ) , one administrator ( Jim Finks ) , and five players ( Warren Moon , Fred Biletnikoff }\\{ \color{red}, John Henry Johnson , Don Maynard , Arnie Weinmeister ) who spent part of their careers in the Canadian Football League ( CFL )}\\{ \color{red} have been inducted ; two of which have been inducted into the Canadian Football Hall of Fame : Warren Moon and Bud Grant.} \\
    \textbf{A2:} {\color{BlueGreen} As of 2018 , 318 individuals have been elected .} \\
    \textbf{A3:} Six players or coaches who spent part of their careers in the short-lived United States Football League ( USFL ) have been inducted . \\
    \textbf{A4:} Current rules of the committee stipulate that between four and eight individuals are selected each year . \\
    \textbf{A5:} Fifteen inductees spent some of their playing career in the All - America Football Conference during the late 1940s . \\
    \bottomrule
    \toprule

    \textbf{WikiQA} \\
    \midrule
    \textbf{Q:} \textbf{how are antibodies used in} \\
    \textbf{A1:} {\color{red} Antibodies are secreted by a type of white blood cell called a plasma cell .}\\
    \textbf{A2:} {\color{BlueGreen} An antibody (Ab), also known as an immunoglobulin (Ig), is a large Y-shaped protein produced by B-cells that is used by the }\\{\color{BlueGreen} immune system to identify and neutralize foreign objects such as bacteria and viruses .}\\
    \textbf{A3:}  {\color{ForestGreen}Using this binding mechanism, an antibody can tag a microbe or an infected cell for attack by other parts of the immune system, }\\{\color{ForestGreen}or can neutralize its target directly (for example, by blocking a part of a microbe that is essential for its invasion and survival).} \\
    \textbf{A4:} Antibodies can occur in two physical forms, a soluble form that is secreted from the cell, and a membrane -bound form that is \\ attached to the surface of a B cell and is referred to as the B cell receptor (BCR). \\
    \textbf{A5:} The BCR is only found on the surface of B cells and facilitates the activation of these cells and their subsequent differentiation into \\ either antibody factories called plasma cells , or memory B cells that will survive in the body and remember that same antigen so the B \\ cells can respond faster upon future exposure. \\
    \bottomrule
    \toprule

    \textbf{TREC-QA} \\
    \midrule
    \textbf{Q:} \textbf{Where is the group Wiggles from ?} \\
    \textbf{A1:} {\color{red} Let 's now give a welcome to the Wiggles , a goofy new import from Australia .}\\
    \textbf{A2:} {\color{BlueGreen} The Wiggles are four effervescent performers from the Sydney area : Anthony Field , Murray Cook , Jeff Fatt and Greg Page .} \\
    \textbf{A3:} In Australia , the Wiggles is like really huge . \\
    \textbf{A4:} His group had kids howling with joy with routines involving Dorothy the Dinosaur , Henry the Octopus and Wags the Dog . \\
    \textbf{A5:} While relatively new to the American scene , the Wiggles seem to be on to something , judging by kids ' reactions to the group 's 
    \\ belly-slapping shows . \\
    \bottomrule

    \end{tabular}}
    \caption{Qualitative examples from AS2 datasets where the baseline RoBERTa-Base model is unable to rank a correct answer for the question at the top position, but our SP pre-trained model can (top ranked {\color{BlueGreen} correct answer by SP}). Here we present the top ranked answers $\{A1,\dots,A5\}$ in the order given by the RoBERTa-Base model. For all these examples we highlight the top ranked answer by the baseline RoBERTa-Base model in {\color{red} red} since it is incorrect, and any other correct answer in {\color{ForestGreen} green}.}
    \label{tab:qualitative}
\end{table*}

%% file: Tables/AS2_only_mlm_table.tex
\begin{table*}[t!]
    \centering
    \resizebox{\linewidth}{!}{
    \begin{tabular}{lccccccccccccccc}
    \toprule
    \multirow{2}{*}{\textbf{Model+ Data Sampling}} & \multicolumn{3}{c}{\textbf{ASNQ}} & &\multicolumn{3}{c}{\textbf{WikiQA}} & & \multicolumn{3}{c}{\textbf{TREC-QA}} & & \multicolumn{3}{c}{\textbf{WQA}} \\
    \cmidrule{2-4} \cmidrule{6-8} \cmidrule{10-12} \cmidrule{14-16}
                           & \textbf{P@1}    & \textbf{MAP}    & \textbf{MRR}   & &\textbf{ P@1}     & \textbf{MAP}     & \textbf{MRR}   & & \textbf{P@1}   & \textbf{MAP}     & \textbf{MRR}  & & \textbf{P@1}    & \textbf{MAP}    & \textbf{MRR}    \\
    \midrule
    
    RoBERTa-Base                                             & 61.8 {\small (0.2)} & 66.9 {\small (0.1)} & 73.1 {\small (0.1)}       & & 78.3 {\small (2.8)} & 85.8 {\small (1.3)} & 87.2 {\small (1.3)}     & & 90.0 {\small (1.9)} & 89.7 {\small (0.7)} & 94.4 {\small (1.1)}      & &  \multicolumn{3}{c}{Baseline}  \\

    \midrule

    + SSP Data (\textbf{MLM-only})   & 63.4 {\small (0.4)} & 67.1 {\small (0.2)} & 73.8 {\small (0.2)}       & & 76.7 {\small (0.9)} & 84.5 {\small (0.7)} & 85.8 {\small (0.7)}     & & 87.4 {\small (1.3)} & 88.8 {\small (0.6)} & 93.1 {\small (1.0)}      & & -0.6\%   & -0.2\%    & -0.3\% \\
    + SSP                            & \bfunder{64.1} {\small (0.3)} & \bfunder{68.1} {\small (0.2)} & \bfunder{74.5} {\small (0.3)}      & & \bfunder{82.9} {\small (0.7)} & \bfunder{88.7} {\small (0.3)} & \bfunder{89.9} {\small (0.4)}          & & \textbf{88.5} {\small (1.2)}  & \textbf{89.3} {\small (0.7)} & \textbf{93.6} {\small (0.6)}                               & & \bfunder{+0.2}\% & \bfunder{+0.6\%} & \bfunder{+0.3}\% \\

    \midrule

    + SP Data (\textbf{MLM-only})    & 62.8 {\small (0.3)} & 67.2 {\small (0.2)} & 73.7 {\small (0.2)}       & & 76.8 {\small (1.6)} & 84.7 {\small (0.8)} & 86.2 {\small (0.7)}     & & 88.8 {\small (1.3)} & 89.8 {\small (0.3)} & 93.7 {\small (0.9)}      & & -1.0\%   & -0.4\%    & -0.6\% \\
    + SP               & \bfunder{64.1} {\small (0.2)} & \bfunder{68.3} {\small (0.1)} & \bfunder{74.5} {\small (0.2)}        & & \bfunder{81.0} {\small (0.8)} & \bfunder{87.7} {\small (0.3)} & \bfunder{88.9} {\small (0.4)}    & & \textbf{90.9} {\small (2.6)} & \textbf{90.1} {\small (0.8)} & \textbf{94.7} {\small (1.3)}                       & & \bfunder{+0.4\%} & \bfunder{+0.7\%}  & \bfunder{+0.5\%} \\

    \midrule

    + PSD Data (\textbf{MLM-only})   & \bfunder{64.1} {\small (0.5)} & 67.3 {\small (0.2)} & 73.7 {\small (0.2)}       & & 79.1 {\small (1.6)} & 85.6 {\small (1.4)} & 87.1 {\small (1.2)}     & & 87.1 {\small (2.8)} & 89.6 {\small (1.0)} & 92.7 {\small (1.3)}      & & -1.3\%   & -0.3\%    & -0.6\% \\
    + PSD                            & 62.6 {\small (0.4)} & \bfunder{67.7} {\small (0.2)} & 73.7 {\small (0.3)}              & & \textbf{80.5} {\small (1.6)} & \textbf{86.4} {\small (1.1)} & \textbf{88.0} {\small (1.0)}                                        & & \bfunder{90.3} {\small (1.3)} & \textbf{90.3} {\small (0.5)} & \bfunder{95.1} {\small (0.7)}              & & \bfunder{+0.4\%}  & \bfunder{+0.7\%} & \bfunder{+0.5\%} \\

    \bottomrule

    \end{tabular}
    }
    \caption{Results (with std. dev. across 5 runs in parentheses) of our pretrained transformer models when fine-tuned on AS2 datasets with MLM-only pre-training. SSP, SP and PSD refer to our pretraining objectives. Results on WQA are relative to RoBERTa baseline. We highlight in bold and underline results like in Table \ref{tab:results_as2}.}
    \label{tab:results_as2_only_mlm}
\end{table*}